\documentclass[letterpaper, 10 pt, conference]{ieeeconf}  

\IEEEoverridecommandlockouts                              

\usepackage{graphics} 
\usepackage{epsfig} 
\usepackage{mathptmx} 
\usepackage{times} 
\usepackage{amsmath} 
\usepackage{amssymb}  
\usepackage{color}
\usepackage[normalem]{ulem}

\title{\LARGE \bf
Towards Decentralized Human-Swarm Interaction by Means of Sequential Hand Gesture Recognition
}

\author{Zahi Kakish, Sritanay Vedartham, and Spring Berman
\thanks{This work was supported by the Arizona State University Global Security Initiative.}
\thanks{Zahi Kakish and Spring Berman are with the School for Engineering of Matter, Transport and Energy, Arizona State University (ASU), Tempe, AZ 85281, USA
      {\tt\small \{zahi.kakish, spring.berman\}@asu.edu}}
\thanks{Sritanay Vedartham is a high school student at BASIS, Scottsdale, AZ 85259, USA
      {\tt\small vedarthamtanay@gmail.com}}
}

\begin{document}

\maketitle
\thispagestyle{empty}
\pagestyle{empty}

\begin{abstract}
In this work, we present preliminary work on a novel method for Human-Swarm Interaction (HSI) that can be used to change the macroscopic behavior of a swarm of robots with decentralized sensing and control. By integrating a small yet capable hand gesture recognition convolutional neural network (CNN) with the next-generation Robot Operating System \emph{ros2}, which enables decentralized implementation of robot software for multi-robot applications, we demonstrate the feasibility of programming a swarm of robots to recognize and respond to a sequence of hand gestures that capable of correspond to different types of swarm behaviors. We test our approach using a sequence of gestures that modifies the target inter-robot distance in a group of three Turtlebot3 Burger robots in order to prevent robot collisions with obstacles. The approach is validated in three different Gazebo simulation environments and in a physical testbed that reproduces one of the simulated environments.

\end{abstract}

\section{INTRODUCTION}

As the price of low-power computing devices continues to decrease, the robotics community recognizes the increasing viability of robotic swarms for a multitude of applications. The emphasis on scalability and hardware redundancy in robotic swarms, however, makes developing robust and stable management and control tools difficult. In addition, a human-in-the-loop approaches for managing a fleet of robots leads to cognitive overload by operators impairing their ability to continue working \cite{chen_humanagent_2014}. Our work seeks to solve this issue in Human-Swarm Interfaces (HSI) by attempting to remove the interface. Instead, a human operator may give commands through a sequence of hand gestures and have their commands relayed to an entire swarm or team of robots.


Yet before this can happen, further progress is required in Human-Swarm interfaces and control. Past work in the field has expanded upon control using a variety of different control interfaces which are often referred to as Human-Swarm Interfaces. Kolling \emph{et al.} provide an essential survey of the HSI field as it currently stands \cite{kolling_human_2016}. As mentioned earlier in \cite{chen_humanagent_2014}, one of the most difficult aspects of human-swarm control is burdening the user with cognitive overload. Managing multiple robots is a difficult task, and, therefore, HSI systems are designed with this premise. Abstracting complex swarm behavior is essential to discretizing the tasks into manageable and comprehensible feedback to the human user. Lin \emph{et al.} were able to generate artificial task functions capable of abstracting a swarm in real-time to assist users \cite{lin_experiments_2015}. After abstracting complex swarm behavior, \cite{ayanian_controlling_2014} and \cite{diaz-mercado_distributed_2015} mapped the different abstracted behaviors onto single gesture inputs for tablet interfaces. However, both these methods require a centralized server to function because of their robot's low computational power, and they require the difficult development of an external mobile application.

Various approaches have been attempted to control multi-robot systems by other physical means, including wearable devices using haptic feedback as seen in \cite{music_human-multi-robot_nodate}\cite{ferrer_wearable_2018} and Electroencephalogram brain-computer interfaces (BCI) as reported by Karavas \emph{et al.} \cite{karavas_hybrid_2017}. However, both these approaches are hindered by the necessity of wearable hardware to function and cumbersome application process to read noisy EEG signals, respectively. These issues make the their applicability in everyday usage more difficult.

In this work, we present preliminary research into a human-swarm control paradigm capable of future use in decentralized control of a robotic swarm through human input that is realizable with the current generation of computer hardware. By decentralized, we mean that the robots' systems are independent of one another but are capable of inter-communication. With the advent of accelerated training in machine learning and advances in computer vision, many complex classification tasks are scalable to less computationally powerful devices. We seek to utilize this by combining advances in decentralized robotic systems using \emph{ros2} \cite{osrf_ros2_2019} with specialized convolutional neural networks (CNN) capable of recognizing hand gestures. We hypothesize that concatenating a series of different hand gestures produces more refined control of a particular swarm task specified by a user. 

 In Section \ref{sec:methodology}, we establish a simplified system composed of a CNN and simplified swarm behavior to validate our hypothesis. Next, we test our approach using a simplified swarm behavior on a small swarm size (three robots) in various simulation testbeds. The system is further validated through physical experimentation using three robots in a re-creation of the final simulated testbed. Finally, we expand upon the results of the experiments and discuss the future applicability of this approach towards a decentralized swarm controlled by a human-in-the-loop.  As of time of writing, this is the only known work that formulates sequential hand gestures for controlling multiple robots using CNNs in a semi-decentralized manner.

\section{METHODOLOGY}
\label{sec:methodology}

Our system is primary composed of two parts: (1) a deep neural network classifying various hand gestures, and (2) a simplified swarm cohesion model for a small swarm size. This paper will provide a validation of sequential hand gestures as a means of controlling a swarm. Therefore, we do not formulate the algorithm as a generalized model for scalable swarm sizes, which we leave to future work.

\subsection{Deep Neural Network}
\label{sec:dnn}

Since low-powered, small computationally powerful devices, such as the Raspberry Pi and other \emph{ARM} devices, are common among swarm robotics hardware architecture, most conventional and state-of-the-art neural network models are unusable or difficult to implement. To circumvent this issue, a special neural network named a SqueezeNet was deployed for hand gesture recognition \cite{iandola_squeezenet_2016}. The SqueezeNet model is capable of performing as well, or even surpassing, classification rates of many the most popular model types with only a fraction of the parameters, thus reducing the model size to roughly $\sim5$ MB. This substantial reduction in model size makes this model ideal for use in hardware typically found in swarm robotics applications.

\subsubsection{Dataset and Preprocessing}
\label{sec:dnn_dataset}

A training set consisting of six different gestures and $2,956$ images was utilized for training the SqueezeNet model \cite{heintz_gestures_2018}. The images are silhouettes of the gestures in black and white. A gesture set, $G$, containing all possible gestures for classification and available in the training data is defined as:

\begin{equation}
    \label{eq:gesture_set}
    G = \{~\textit{C}, ~\textit{Fist}, ~\textit{L}, ~\textit{Ok},~\textit{Peace}, 
         ~\textit{Palm}~ \}
\end{equation}

\begin{figure}[!t]
	\centering
	\includegraphics[width=0.45\textwidth]{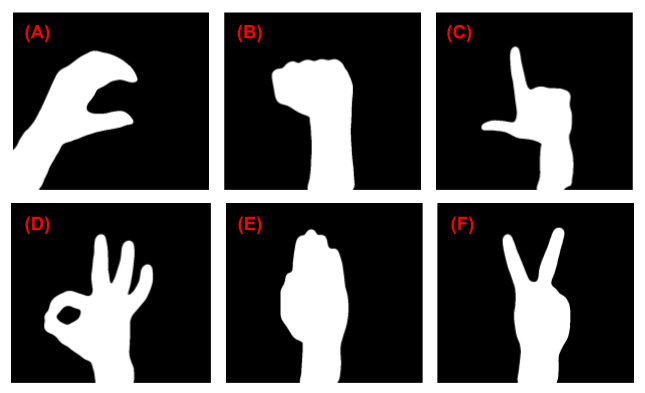}
	\caption{Silhouettes of gestures available through the dataset in \cite{heintz_gestures_2018}. In addition to these gestures, a blank image was used to classify the gesture \emph{None}. A) a \emph{C} shape B) a \emph{Fist} C) an \emph{L} shape D) the \emph{Okay} sign E) a \emph{Palm} F) the \emph{Peace} sign.}
	\label{fig:silhouttes}
\end{figure}

Figure \ref{fig:silhouttes} contains examples of the gestures found in the dataset. The dataset was expanded to ensure more robust classification of gestures during real-time operation. The data augmentation began with inverting the images horizontally to resemble the gesture but with the opposite hand. The original image and the inverted image are used to generate three new images, respectively. One rotated by $90^{\circ}$ clockwise, another $90^{\circ}$ counter-clockwise, and an image flipped upside-down. Therefore, seven new images are generated for each image in the original dataset.

To ensure that the model is capable of real-time performance, a seventh gesture was added to the dataset: `None'. Since the user's hand is not consistently in the frame, the image is classified as `None' and awaits a gesture to come into the frame. A dataset containing blank images is generated during preprocessing consisting of black images and black images containing Gaussian noise to add to SqueezeNet model's classification capabilities.

Further preprocessing was performed before training to ensure optimal results. Images in the dataset were cropped from their original size of $640 \times 576$ to $570 \times 570$. The images were then scaled to $240 \times 240$. This final image size is big enough to ensure adequate real-time classification of gestures without increasing computational cost.

\subsubsection{Training}
\label{sec:dnn_training}

The SqueezeNet model was programmed and trained using the \emph{TensorFlow} API version $1.14$ \cite{tensorflow_2015} running on a development box containing four \emph{Nvidia} RTX 2080 graphics cards. The model was trained for $10$ epochs with a Stochastic Gradient Descent optimizer, which had a learning rate $\alpha = 0.001$, a coefficient of momentum $\eta = 0.9$, and a learning rate decay $\lambda = 0.0002$. The batch size was set to two images to prevent memory issues due to the images' size. The dataset was split into a training and validation set consisting of $90\%$ and $10\%$ of the images, respectively.

\subsection{Gesture Driven Controller}
\label{sec:controller}

We first consider a waypoint system in which a finite number of robots, $N$, plot a straight motion to a desired goal position $(x_{d}, y_{d})_{N}$ from an initial starting position $(x_{i}, y_{i})_{N}$. To help validate the applicability of sequential gesture control, we will only focus on one direction, $x$, and keep the $y$ values arbitrary. For example, if the initial position of a robot is $(0, 0)$ and the desired position is set to be $(10, 0)$, the desired position is considered achieved if only the robot's x-coordinate matches the desired value. A set of waypoints, $\mathcal{W}$, for each robot $n$ is thus generated from augmenting $x_i$ by $x_\text{offset}$, which is calculated from a user-defined number of waypoints, $M$.
\begin{align}
    x_{\text{offset}} &= \frac{|x_{i} - x_{d}|}{M} \label{eq:somethign} \\
    \mathcal{W}_{n} &= \lbrace (x_{i} + k \times x_{\text{offset}}, y): k = 0,\ldots,M \rbrace \label{eq:waypoins}
\end{align}
Each robot, $n$, moves from one waypoint to another in the set $\mathcal{W}$ until it reaches the final desired position $(x_{d}, y_{d})_{n}$.

With the waypoints generated, we define $\mathcal{C}: y_{n} \times \beta \to \mathbb{R}$ to be the cohesion of the swarm, that is, a metric measuring the inter-robot distance of a swarm, where $y_n$ is the robot $n$'s $y$ position and $\beta$ is the cohesion factor. $\beta$ is a binary variable that represents an increase in cohesion if $\beta = 0$ or a decrease in cohesion if $\beta = 1$. To increase the cohesion of a swarm is to reduce the distance between each robot, and to decrease the cohesion is to increase the distance between each robot. Additionally, we consider a small swarm size of three agents in a line separated by an offset in the $y$ direction. The cohesion of this small swarm is best represented by a constant addition or subtraction of a small, static offset $y_{\text{offset}}$. An individual robot, $n$, applies a cohesion change by the following piecewise formula:

\begin{equation}
    \label{eq:piecewise}
    \mathcal{C}(y_{n}, \beta) =
    \begin{cases}
      y_{n} - y_{\text{offset}}  & \quad \text{if } y_{n} > 0 \text{ and } \beta = 0 \\
      y_{n} + y_{\text{offset}}  & \quad \text{if } y_{n} > 0 \text{ and } \beta = 1\\
      y_{n} + y_{\text{offset}}  & \quad \text{if } y_{n} < 0 \text{ and } \beta = 0\\
      y_{n} - y_{\text{offset}}  & \quad \text{if } y_{n} < 0 \text{ and } \beta = 1\\
    \end{cases}
\end{equation}

Next, we consider a differential drive capable of driving from one waypoint to the next based on the unicycle model \cite{carona_control_2008} as defined by the following state-space formulation, where $x$ and $y$ are the robot's position and $\phi$ is its angle.
\begin{equation}
    \label{eq:unicycle}
    \dot{x} =
    \begin{bmatrix}
    \dot{x}    \\
    \dot{y}    \\
    \dot{\phi} \\
    \end{bmatrix}
    =
    \begin{bmatrix}
    v_{o} \cos \phi \\
    v_{o} \sin \phi \\
    \omega
    \end{bmatrix}
\end{equation}
Here, $v_{o}$ and $\omega$ is the robot's linear and angular velocity, respectively. Since we choose to apply a constant linear velocity, the dynamics of the robot using the unicycle model are represented primarily by $\omega$. To control the robot's motion from one point to another reference point, the difference between a desired angle, $\phi_d$, and the robot's current angle, $\phi$, is calculated and used as the error
\begin{equation}
	\label{eq:error}
    e = \phi_d - \phi.
\end{equation}
To prevent singularities from arising and restrict the error between $0$ and $2\pi$, we update the calculated error using $\mathit{atan2}$

\begin{equation}
    \label{eq:error_new}
    e_{new} = \mathit{atan2}(\sin(e), \cos(e)).
\end{equation}

Finally, our control input into the robot is applied with a proportional gain $K_{p}$.

\begin{equation}
    \label{eq:controller}
    \omega = K_{p} \dot e_{new}
\end{equation}

This forms the basis for the control of the swarm using sequential hand gestures.

\section{SIMULATIONS}
\label{sec:simulation}

A series of 3D simulations were designed to test the viability of sequential gesture-based control of a swarm. To formalize the experimental simulations, the following assumptions were made. First, the simulations are limited to three robots. Though it may seem like a small swarm size, we believe that this size provides a viable minimal benchmark for how our algorithm will actually work. Second, odometry is provided by the simulation environment and not by internal (i.e. encoders) or external (i.e. GPS, cameras, etc.) sensors. Finally, a black background is used when reading images from the gestures to streamline classifications. 

\subsection{Structure}
\label{sec:sim_structure}

The simulation was developed using \emph{ros2}, the next-generation of the Robot Operating System (ROS). Compared to the original ROS, ros2 provides an enhanced middleware programming environment, the removal of a ROS Master to make multi-robot decentralized approaches easier, and expanded platform and architecture support. \emph{Gazebo}, a 3D robot simulation environment, is used to render the experiment. Gazebo is a project developed in tandem with ROS/ros2, and comes with many ROS drivers to simplify development of robots by providing a simulation environment that functions similarly to a physical one. The experiment is run using the Turtlebot3 Burger robot by \emph{ROBOTIS} \cite{robotis_tb3_2019} who provide 3D Gazebo models for use in simulation.

Each robot runs two separate nodes: one that contains drivers to connect to Gazebo to simulate differential drive control, sensor readings, etc., and another that is the primary motion controller as defined in Section \ref{sec:controller}. The robots do not intercommunicate and only subscribe to messages from one external node that tells them what gestures were classified. This external node reads information from a generic webcam running the SqueezeNet model. Additionally, images captured by the webcam are modified to reflect those used for training, which were silhouette images of the different hand gestures.

The captured camera images are cropped from the $640\times480$ resolution to $480\times480$ resolution and then scaled to $240\times240$, which is the input size for our SqueezeNet model. The images are then converted to greyscale and a slight Gaussian blur is applied to prevent fine details in the image from causing artifacts when converted to a binary image. Finally, the image is converted using a binary filter and inputted into the SqueezeNet model. The predicted gesture is then published to all the robots. This message is not unique to each robot.

\subsection{Gestures}
\label{sec:sim_gestures}

For this experiment, the number of gestures used have been reduced to the following five:

\begin{equation}
    \label{eq:actual_gestures}
    G_{possible} = \{\textit{Palm}, ~\textit{Peace}, ~\textit{Fist}, ~\textit{C},
                    ~\textit{L}\}
\end{equation}

These gestures are mapped to the subsequent actions the swarm may undergo:

\begin{itemize}
    \item \textbf{Palm}: Stop movement of the swarm
    \item \textbf{Peace}: Resume movement of the swarm
    \item \textbf{Fist}: Read cohesion action
    \item \textbf{C Sign}: Increase
    \item \textbf{L Sign}: Decrease
\end{itemize}

Additionally, there is a sixth classified gesture is \emph{None}, which, as stated in Section \ref{sec:dnn_dataset}, means that there is no gesture recognized and no swarm action applicable. \emph{Palm} (Stop) and \emph{Peace} (Resume) are the only two gestures capable of controlling the swarm alone. The rest require the user to give a sequence of gestures for the swarm to read. One a gesture pertaining to the swarm behavior (cohesion) the user wishes to modify and the next is a gesture mapped to a modification variable (increase or decrease).

In the simulation, we will rely on two sequences used to modify the cohesion of the swarm to help negotiate an obstacle. The first is increasing the cohesion of the swarm, $\beta = 0$, which means that the swarm will group closer to one another. This is done by giving the following commands in this order:

\begin{equation}
    \textit{Palm} \rightarrow \textit{Fist} \rightarrow \textit{C}
    \rightarrow \textit{Peace}
\end{equation}

This sequence is explained as \textit{``stop the swarm, read my cohesion command, increase cohesion by one step size, and resume moving."} The second would be to decrease the cohesion of the swarm, $\beta = 1$, resulting in a increase in distance between the robots. This is done using the same hand gestures but with the decrease cohesion command.

\begin{equation}
    \textit{Palm} \rightarrow \textit{Fist} \rightarrow \textit{L}
    \rightarrow \textit{Peace}
\end{equation}

Just like the increase cohesion sequence, this sequence is similarly explained as \textit{``stop the swarm, read my cohesion command, decrease cohesion by one step size, and resume moving.''} As described earlier, the cohesion of the swarm is set as steps. Specifically, each call to increase or decrease the swarm cohesion decreases or increases the distance between the robots by a calculated $y_{\text{offset}}$, respectively. If an obstacle requires the user to change the swarms cohesion by multiple steps of $y_{\text{offset}}$, the user does not need to repeat the whole sequence twice but can concatenate the swarm command multiple times. For example, if the swarm cohesion is needed to increase by two steps, a user would provide the following command:

\begin{equation}
    \textit{Palm} \rightarrow \textit{Fist} \rightarrow \textit{C}
    \rightarrow \textit{Fist} \rightarrow \textit{C}
    \rightarrow \textit{Peace}
\end{equation}

\begin{figure}[!t]
	\centering
	\includegraphics[width=0.45\textwidth]{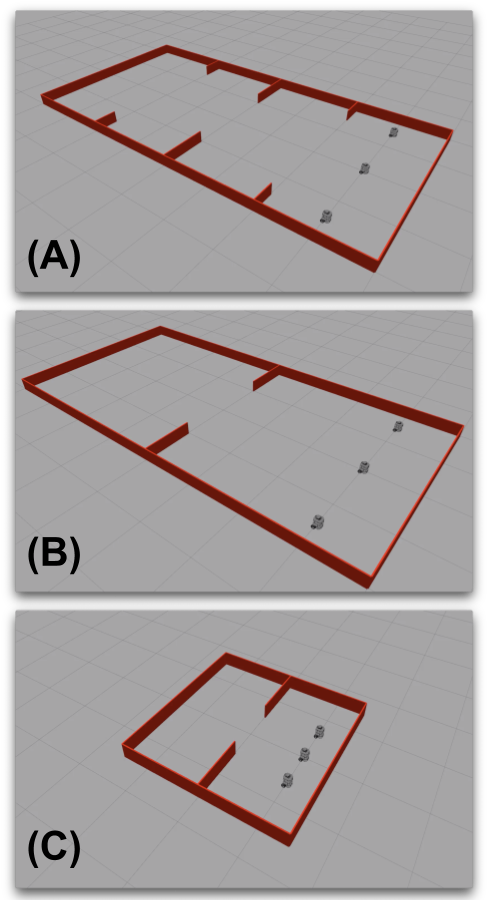}
	\caption{The three testbeds created for simulation. A) An $8m\times4m$ testbed containing multiple openings for the swarm to traverse through. B) An $8m\times4m$ testbed containing only one small opening for the agents to negotiate. Compared to the previous testbed, this requires the user to string together multiple sequences of the same gesture to complete. C) A small $2.5m\times2.5m$ testbed with one small $1m$ wide opening. This last testbed is recreated for physical validation in Section \ref{sec:pe}}
	\label{fig:testbeds}
\end{figure}

\subsection{Simulated Environment}
\label{sec:sim_exp_design}

The three robots are placed in a line within three simulated testbeds, shown in Figure \ref{fig:testbeds}. The first contains a series of two types of openings: one small-sized opening in the middle and two intermediate-sized openings located on both ends of the testbed. The second testbed has only one small-sized opening. Each testbed provides a different validation for the capability of our sequential gesture control scheme. The first demonstrates how individual commands can be given at the onset of an obstacle, and the other shows how for more difficult obstacles a user is capable of stringing together multiple gesture actions into one command. The first two testbeds are $8m \times 4m$ in size. Each robot is placed $1m$ from the end of the testbed and spread to have an inter-robot distance of $1.5m$ between one another. To complete each task, the robots will have to move forward and reach the other end of the testbed. The swarm of robots will need to negotiate the obstacles before them by relying on a user's sequential gesture input.

\begin{figure*}
	\centering
	\includegraphics[width=0.8\textwidth]{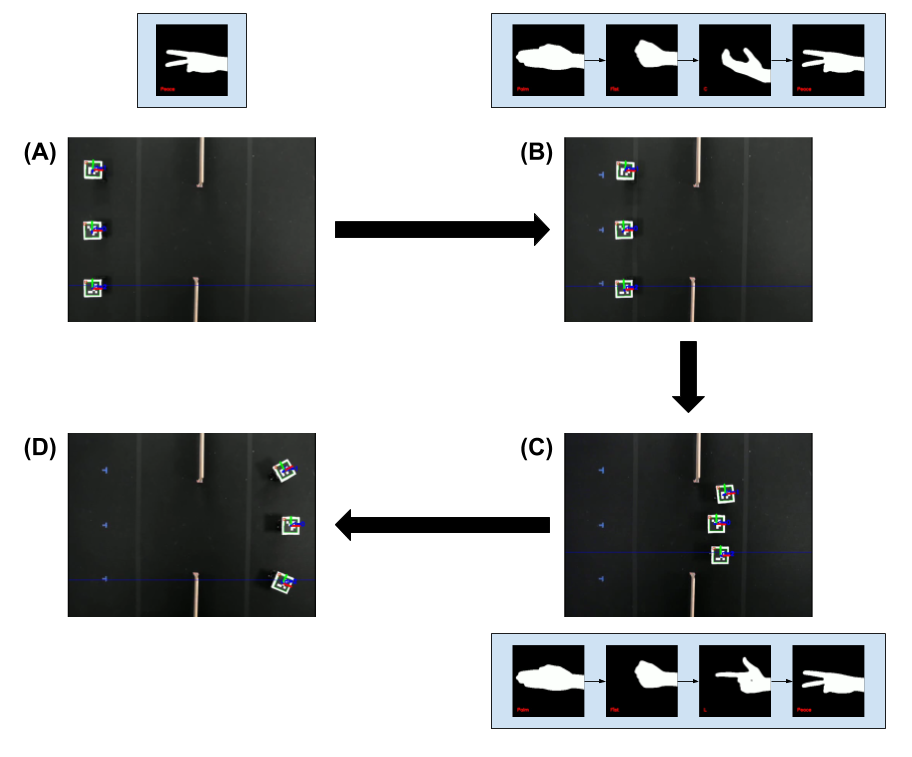}
	\caption{An overview of the physical experiment with gesture sequences required for the swarm to get through the small opening in the middle of the arena. A) The robot swarm begins on one end of the testbed. The \emph{Peace} gesture is used as a standalone command to start the experiment. B) Once the robots get closer to the obstacle, the following sequence ($\textit{Palm}\rightarrow\textit{Fist}\rightarrow\textit{C}\rightarrow\textit{Peace}$) is given to increase the cohesion of the swarm and resume motion. C) After the robots have cleared the opening, another gesture sequence ($\textit{Palm}\rightarrow\textit{Fist}\rightarrow\textit{L}\rightarrow\textit{Peace}$) is used to return the swarm back to their initial cohesion size. D) Finally, the robots reach the final waypoint and the experiment completes.}
	\label{fig:gesture_diagram}
\end{figure*}

The third testbed is a recreation of a real testbed used in the physical experiment section of the paper. Compared to the large surface area of the other testbeds, this testbed is significantly smaller at $2.5m \times 2.5m$. Additionally, the initial inter-robot distance is reduced to $0.6m$ and the robots are placed $0.5m$ from the end. The testbed contains a $1m$ sized opening a meter from the robots starting position. Like the other testbeds, the robots will have to negotiate this obstacle using a single sequence of gestures to increase cohesion. After going through the obstacle, a different sequence is used to decrease the cohesion. This testbed's results act as validation for the success of this experiment using real robots. Figure \ref{fig:gesture_diagram} provides a brief overview of the robots increasing and then decreasing their cohesion to negotiate a small opening.

\section{PHYSICAL EXPERIMENTS}
\label{sec:pe}

\begin{figure}[!t]
	\centering
	\includegraphics[width=0.4\textwidth]{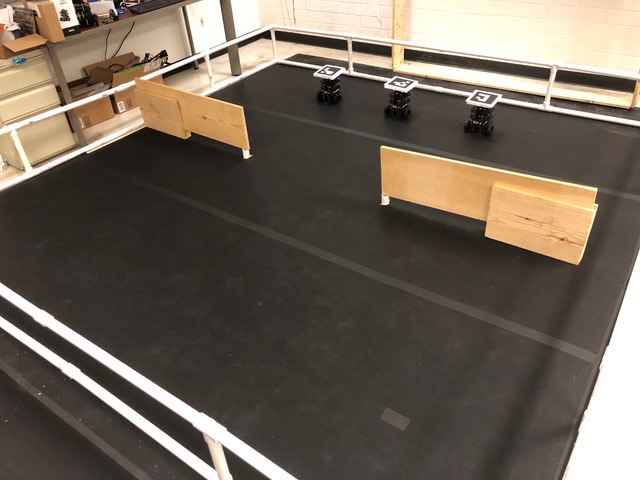}
	\caption{The physical testbed.}
	\label{fig:phy_testbeds}
\end{figure}

To validate the use of sequential hand gesture control of a swarm in a physical setting, an analogous physical testbed to the third simulated testbed was built as shown in Figure \ref{fig:phy_testbeds}. The experiment was set up in the same manner as the simulated one. For this paper, the individual robot controller and gesture recognition code was written with the intention of interchangeability between the simulated and physical experiments. The only difference were parameter files which provided environmental constraints and individual robot attributes depending on the environment the test is being run.

However, certain aspects of the simulation were not available for use in our physical experiments. Odometry within simulation is accurately calculated by the Gazebo environment, but getting this same information in a physical experiment required use of an overhead camera and \emph{ArUco} fiducial markers \cite{ramirez_aruco_2018} \cite{garrido_aruco_2015} to calculate each individual robot's pose. A ros2 node was developed to track individual robot position from the overhead camera and calculate poses from a $0.127\textit{m} \times 0.127\textit{m}$ marker placed atop each robot that was detected. This odometry information is then published to the robots for use in their controller node's.

During simulation, all the ros2 individual robot nodes and gesture node were run on the same computer. The physical experiment distributes the computation over a wireless network. Each robot runs their respective controller on their hardware, but do not communicate to one another. The only communication they receive are from two external sources: the node calculating individual robot odometry from a central overhead camera and the node reading and classifying the sequential gestures from the user. Each of these nodes are also run on separate computers due to convenience rather than the inability of running both on the same one. This decentralization comes with a cost, however, in the form an approximately $0.5s$ delay. Even though the individual robots were capable of running the gesture detection node, we chose to keep the node running on a separate computer to keep the test runs between the simulation and physical experiments similar.

The physical experimental procedure is nearly identical to the simulation except for the software structure changes presented in this section. There are a few small changes in comparison to the simulation; however, we do not believe it reduces the validity of our physical experiment. One additional change to this experiment was the linear velocity of each robot was reduced due to the half-second network lag in the system.

\section{DISCUSSION}
\label{sec:discussion}

We have successfully demonstrated our hypothesis by showcasing a simplified cohesion control model for both simulated and physical testbeds. The supplemental video attached provides the results of a single run on each simulated and physical testbed. Each simulated test completed successfully and the controller responded correctly to the properly classified sequential gesture commands given from the SqueezeNet model in real-time. As mentioned in Section \ref{sec:sim_structure}, the run corresponding to Testbed $2$ demonstrated the ability for the systems to read multiple instances of the same \emph{increase cohesion} gesture sequence in one input. Results from that run show that the provided input sequence was easily registered and enacted by the robots. Additionally, the physical experiment was able to finish successfully even with the network delay present in the system. We believe that re-creations of bigger testbeds, such as Testbed $1$ and $2$ in our simulations, would yield successful runs. Although the system is semi-decentralized due to odometry calculations and having the gesture recognition node running separate from the robots, these results of these tests show the feasibility of a human operator interacting with a decentralized robot swarm by showing a robot a sequence of hand gestures.

Although all the experiments were successful, we did run into a minor classification issue during test runs. The SqueezeNet CNN would sporadically misclassify the hand gesture upon the subject's hand leaving the camera's viewing area or when switching between hand gestures. We believe that this issue is likely due to gestures created during the hand's motion unable to be classified. To help reduce this error, publishing of the predicted gesture was limited to once every half second instead of one every tenth of a second, which is the refresh rate of the ros2 node that classified the gestures.

\section{FUTURE WORK}
\label{sec:future_work}

As stated in the beginning of Section \ref{sec:methodology}, the experiments shown were not generalized, but future work hopes to encompass multiple, generalized swarm behaviors. Research by Harriott \emph{et al.} gives an interesting breakdown of numerous biologically-inspired swarm behavior metrics \cite{harriott_biologically-inspired_2014} as possible future behaviors employable by our system. A further expanded repertoire of behaviors would increase the feasibility of this system for everyday work with human operators.

Although this light-weight SqueezeNet model is capable of being run on low-powered hardware such as the Raspberry Pi due to its size (approximately $\sim5$ MB), the model proved very sensitive to lighting conditions and cluttered backgrounds. Generating the silhouette images is very difficult in more dynamic backgrounds, which we believe is necessary if a gesture recognition system were to be placed on each robot. Therefore, all of our tests were run with a static black background. As of time of writing, the team at \emph{Google AI} has developed algorithms for more robust hand gesture recognition without the need for silhouette images capable of running on Android and iOS devices \cite{bazarevsky_google-ai_2019}. These algorithms are capable of running on hardware similar to that of the Turtlebot3 and is able to classify a larger set of hand gestures. Our future work seeks to explore this and other algorithms capable of recognizing hand gestures in more dynamic and cluttered backgrounds.

Finally, we intend this work to help us pursue human-swarm interaction capable of rewarding a swarm for the actions or tasks it accomplishes. Specifically, a form of dynamic on-line reinforcement learning (RL) on a decentralized swarm. Work by Nam {et al.} show the feasibility in applying RL to predict or learn a human operators ``trust" is in the swarm they are operating \cite{nam_models_2019} \cite{nam_predicting_2017}. We hope that future work allows more unique and novel methods for human and swarm to learn from one another in different environments.



\section{CONCLUSION}
\label{sec:conclusion}

Our paper provides preliminary work into using a sequence of hand gestures to control swarm behavior. This system is a combination of a small sized CNN model capable of recognizing silhouette images of hand gestures in real-time and a decentralized robot development environment, \emph{ros2}. We test this control strategy in a semi-decentralized manner in three simulated testbeds using three Turtlebot3 Burger robots. The system is further tested on a physical testbed with the same robots. Both environments yielded successful runs using a single swarm metric (cohesion of the swarm). In the future, we intend to expand upon this work and create a fully realized, decentralized swarm system controllable by a human operator with no external equipment by giving a sequence of commands to control the whole swarm.




\section*{ACKNOWLEDGMENT}

The authors would like to thank Shiba Biswal (ASU), Karthik Elamvazhuthi (UCLA), and Varun Nalam (ASU) for their assistance in this work.


\bibliographystyle{unsrt}  
\bibliography{icra2020}

\end{document}